# Applications of Machine Learning in Chemical and Biological Oceanography

Balamurugan Sadaiappan, Preethiya Balakrishnan, Vishal C.R., Neethu T. Vijayan, Mahendran Subramanian, and Mangesh U. Gauns*

**ABSTRACT:** Machine learning (ML) refers to computer algorithms that predict a meaningful output or categorize complex systems based on a large amount of data. ML is applied in various areas including natural science, engineering, space exploration, and even gaming development. This review focuses on the use of machine learning in the field of chemical and biological oceanography. In the prediction of global fixed nitrogen levels, partial carbon dioxide pressure, and other chemical properties, the application of ML is a promising tool. Machine learning is also utilized in the field of biological oceanography to detect planktonic forms from various images (i.e., microscopy, FlowCAM, and video recorders), spectrometers, and other signal processing techniques. Moreover, ML successfully classified the mammals using their acoustics, detecting endangered mammalian and fish species in a specific environment. Most importantly, using environmental data, the ML proved to be an effective method for predicting hypoxic conditions and harmful algal bloom events, an essential measurement in terms of environmental monitoring. Furthermore, machine learning was used to construct a number of databases for various species that will be useful to other researchers, and the creation of new algorithms will help the marine research community better comprehend the chemistry and biology of the ocean.

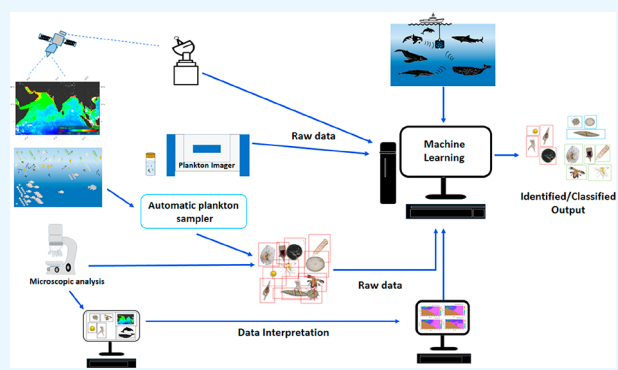

## 1. INTRODUCTION

The ocean encloses complex ecosystems, each with a distinct physical, chemical, and geological composition, supporting a vast spectrum of species. As the ocean covers 71% of the earth's surface, it supports more living organisms than the terrestrial habitats.[1] The earth's biogeochemical cycle is heavily reliant on the ocean. Furthermore, it absorbs more than half of the carbon in the atmosphere. It also serves as the primary oxygen source. Due to its vast nature and complicated environment, continual monitoring is required to fully comprehend the ecosystem. As modern science progressed, additional research in the ocean environment was conducted on both a local and global scale. Even the most distant sections of the Antarctic and Arctic regions are monitored. Decades of investigations yield a huge amount of data that reflect these ecosystem characteristics. It does become increasingly difficult to analyze big data using conventional numerical approaches.

Biological oceanography deals with understanding physical, chemical, and other oceanography processes that influence the distribution and abundance of various types of marine life and additionally deals with how living organisms like viruses, microbes, plankton, and animals behave and interact with biogeochemical processes in the oceans.[2] Furthermore, it deals with how species adapt to environmental changes like the rise in temperature and pollution. Among the living organisms in the ocean, phytoplankton is the primary producer and plays an important role in the food web, as well as the biogeochemical cycle in the ocean,[3] and acts as an indicator of pollution or eutrophication or climatic events (change in abundance and distribution).[4] Similarly, zooplankton play an important role in the marine food web, elementary cycle, and vertical fluxes. Likewise, different eukaryotic organisms thrive in the ocean including fishes, turtles, dolphins, sharks, and mammals like whales, etc. All of these organisms play a role in their ecosystem but are collectively facing difficulties due to climate change. Chemical oceanography demonstrates the spatial and temporal distribution of elements, molecules, atoms, and compounds, which are closely related to biological, physical, and geological oceanography. It is involved in the study of carbon, nitrogen, sulfur, and other element cycles. The ocean holds a larger portion of inorganic carbon than the atmosphere. The flow of chemicals in the ocean, especially carbon, depends on the earth's climate. The atmospheric $CO_2$ concentration increases rapidly due to the use of fossil fuel, which in turn increases the ocean carbon level.[5]

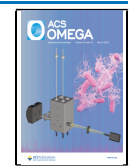

The discipline of Machine Learning (ML) is concerned with the use of computer algorithms to solve issues. It is a potential approach for allowing computers to assist people in analyzing huge and complicated data sets. ML is a type of statistical computing in which computers anticipate the outcome based on the input data.[6] Furthermore, it is a subject that deals with constructing models, performing analysis, classification, and prediction based on existing data. ML techniques are frequently used to solve a range of large data issues, including image identification and classification and extreme events in complex systems. In a complicated system, predicting and understanding unanticipated statistics is a difficult challenge.[7] ML is the development of dynamic algorithms that can make data-driven choices. Because it can create a model from highly dimensional and nonlinear data with complicated relations and missing values, it also has an edge over the traditional technique. In many aspects of the Earth system (land, ocean, and atmosphere), machine learning has proved to be quite beneficial.[8]

ML is divided into several categories, including supervised, unsupervised, ensemble techniques, neural networks, deep learning, and reinforcement learning. In supervised learning, algorithms require an external supervisor called training data. Human assistance usually provides precise input to get the aimed output for prediction accuracy in training algorithms. In this, the algorithm first obtains knowledge from the training data set to predict and classify the other data. In unsupervised learning algorithms, the computer program automatically searches for a feature or pattern from the given data and groups them into clusters without explicit programming. It uses the previously learned features to classify the new data. It does not need a supervisor or assistance for predicting the given data, so it is called unsupervised learning. Ensemble models combine results from different models. It is an ensemble or collection of decision trees. It is also a versatile algorithm capable of performing both regression and classification. e.g., random forest (RF).

Deep learning models are capable of focusing on the correct features on their own and need little guidance. These models also partly fix the issue of dimensionality. The concept behind deep learning is to construct learning algorithms that mimic the human brain. Deep learning is a group of statistical systems that gain knowledge of strategies used to examine characteristic hierarchies, predominantly based on Artificial Neural Networks (ANNs). It has an input layer, a hidden layer, and an output layer. Such systems discover ways to make predictions via thinking about examples, typically without challenging explicit programming. Back-propagation is a popular deep learning technique for performing supervised multilayer perceptron training. Convolutional neural networks (CNN) consist of a sequence of layers like convolutional (input layer), pooling (reduce dimensions), and fully connected layer (classification). CNN is a sort of feed-forward ANN in which the connectivity patterns between its neurons are influenced by the visual cortex organization of the animal. A computer understands an image or data and uses the layer to process the final class score. In Reinforcement learning (RL) algorithms, the learning is based upon trial-and-error methods. RL uses various software to find the best behavior or result. The decision is made based on action taken, which gives more positive results.

Here, we focused on the application of ML approaches to understand the big data associated with chemical and biological oceanography. Many researchers have utilized ML algorithms to address oceanography-related issues, such as determining phytoplankton dynamics, oceans remote sensing, habitat modeling and distribution, species identification, ocean monitoring, and resource management.

## 2. CHEMICAL OCEANOGRAPHY

The ocean's carbon, micronutrients, and macronutrients are controlled by the physical, chemical, biological, and geological processes, driving the worldwide ocean biogeochemical cycles.[9] Changes in ocean biogeochemical processes in one place may have an effect on the global stage. The recent climate change issue, notably the rise in global temperature, has a significant impact on the biogeochemical cycle. As a result, getting a greater knowledge of the elements that drive change and measuring the impact is a difficult, but necessary, task ahead. In this regard, we have discussed a few studies that employed machine learning to anticipate or estimate the elements in the ocean.

The Gaussian Mixture Model (GMM), an unsupervised ML classifier, was used to understand the spatial variability of physical and biogeochemical properties in the intermediate and deep waters of Southern Ocean (SO).[10] The ML, trained with Argo-based data (temperature and salinity) from 300 to 900 m depth, not only predicts the location and boundaries of the frontal zones but also organizes them into five frontal zones. Moreover, the model predicts the water mass property variations relative to the zonal mean state. The ML model also showed the variability is property dependent and may be twice as intense as the mean zone variability in intense eddy fields. Also, the ML model showed the intense variability in the intermediate and deep waters of the Subtropical Zone; in the Subantarctic Polar Frontal Zone, it was closely related to the intense eddy variability that enhanced the convergence and mixing of the deep water with surface water.

**2.1. Dissolved Oxygen (DO).** The concentration of DO in the ocean affects a variety of factors, including seawater quality, global temperature control, ecosystems, biogeochemical cycling, ocean ventilation, and internal ocean circulation. The mixing of atmosphere-ocean interaction and a net amount of respiration of organic matter in the water column control the ocean $O_2$ concentration.[11] Over the last century due to anthropogenic activities including excess fossil fuel use, the amount of the $O_2$ concentration in the coastal and open ocean waters decreased.[12] But there is no adequate data to find the seasonal and interannual variability on a global scale. In this section, we have discussed a few studies that employed the ML approach to assessing DO concentrations in various ocean realms.

Strong air-sea fluxes and oceanic instabilities are distinct features of the SO. The melting of polar ice caps due to global climate change has also affected the SO. As a result, research in these areas is of global importance. The DO concentration at 150 m depth of the SO was accurately estimated using Random Forest Regression (RFR) with given temperature, salinity, location, and time.[13] On validation with synthetic data from the Southern Ocean State Estimate (SOSE), the RFR model performed well in estimating the concentration of $O_2$ in most regions, while some boundary regions were difficult to predict. Additionally, RFR predicted that both the SOSE and World Ocean Atlas (WOA13) overestimate the yearly mean $O_2$ (at 150 m depth) in the SO, both a global and basin scale; the model predicts that the SOSE may underestimate the annual cycle. The model also predicted a large regional bias in the east of Argentina. Overall, the RFR proved to be a better tool for understanding annual mean $O_2$ and variability from the other sparse $O_2$ measurements. This RFR model may be useful to map other biogeochemical variables.

ML has shown to have better performance in the prediction of DO. However, the combination of different ML algorithms has its advantages, a combination of tree-based models and neural networks termed a Marine-Deep Jointly Informed Neural Network (M-DJINN) estimates the DO in the ocean.[14] The M-DJINN method uses a zero-mean Gaussian distribution to predict the marine DO concentration. By choosing the number of trees and the maximum depth of trees, the M-DJINN proved to be more efficient than DJINN in terms of computing time and prediction ability. M-DJINN performed better in terms of accuracy and convergence in predicting marine dissolved oxygen from the World Ocean Database 2013 (WOD13) data set. For this prediction a random oceanographic data set from WOD13 for the years 2001 to 2010 was used; the data set includes temperature (T), salinity (S), phosphate (P), and DO. Apart from the calculation time, M-DJINN decreases the mean squared error in predicting oxygen concentration to 17.6% (when the max tree depth was fixed at 10).

DO in the coastal regions is as important as DO in the SO, as it supports the coastal economy. Apart from predicting the DO in water bodies, it is equally important to predict hypoxic conditions well before they happen. Hypoxia, a low concentration of DO (less than 2 mg/L) in water bodies, mostly occurs in estuaries and coastal waters. One of the major factors that causes mortality in fishes and other aquatic organisms, in turn, alters the ecosystem community and influences the biogeochemical cycles.[15] Recent climate change and altered physical conditions aggravate hypoxia. So, early prediction of these conditions is essential for environmental management. However, the accurate prediction of DO spatial-temporal variation and hypoxia is still a difficult task, even with advanced numerical methods. ML models like RFR and Support vector regression (SVR) with little training data sets accurately predicted the offshore and nearshore DO concentration.[16] Both the models with measured DO concentration from offshore, nearshore, and measured input parameters accurately reproduced the DO concentration. Among the models, RFR performed better than SVR (difficult to tune and took a longer training time). The model showed high accuracy in predicting the DO value with training data from the same site but performed moderately in predicting the DO value at one site with training data from another site. The model also has some abilities like correcting the missing data in time series data sets and detecting coastal hypoxic conditions directly or indirectly. Future iterations of such ML models may produce an accurate real-time forecast of hypoxic events.[16] Apart from the simple ML algorithms, neural networks were also used to estimate or predict the DO concentration in the coastal environment. The spatial-temporal variations of DO and hypoxic conditions in the Chesapeake Bay, USA, were predicted using a neural network[17] where the data were processed in three major steps, i.e., empirical orthogonal functions analysis, automatic selection of forcing transformation, and a neural network. The model has high accuracy with external forcing as model input rather than the in situ measurements. This model proved to be useful in coastal systems that are systematically monitored.

Even with different machine learning algorithms, the accurate forecasting of DO is still challenging. The nonstationary and extreme volatility nature of the DO makes it difficult to predict. Even with predictors and applying different ensemble models, the accurate forecasting of DO is a challenging task. The complexity of using multiple factors affecting DO and applying different ensemble models were overcome by using the gray relational (GR) degree method, empirical wavelet transform (EWT), and multimodel optimization ensemble optimization ensemble. Among the models, a novel hybrid model MF-RNNs-EWT-BEGOE based on the weightage obtained by particle swarm optimization and gravitational search algorithm had better prediction accuracy.[18] The model was shown to be superior with excellent accuracy in forecasting DO; also the model has the ability to predict the trend and enable humans to have better management decisions.

**2.2. Carbon (C).** The oceanic uptake of $CO_2$ caused ocean acidification (increased by ∼0.1 pH units), which may lead to biodiversity loss.[19] The ocean's role in the carbon cycle and spatial heterogeneity of the $CO_2$ flux can be determined by measuring the sea-surface partial pressure of carbon dioxide ($pCO_2$), an essential parameter in quantifying air−sea $CO_2$ flux. Thus, it is important to understand the oceanic uptake and dynamics of the $pCO_2$. The $pCO_2$ was estimated by shipboard (in situ), Agro floats, and satellite image data sets. Satellite data-based estimation was found to be a promising field that required less time and cost. The practical difficulty in measuring $pCO_2$ from satellite data is that multiple environmental factors control the $pCO_2$. Earlier, $pCO_2$ was estimated using regression and multiple regression[20−27] from the satellite data like sea surface temperature (SST), sea surface salinity (SSS), chlorophyll-a (Chl-a) concentration, downwelling of irradiance ($K_d$), wind, and mixed layer depth (MLD). These analyses were not able to accurately estimate $pCO_2$ in large oceanic regions. Also, features (predictors) play an important role in determining $pCO_2$, so selecting parameters from the satellite data for estimating $pCO_2$ was important. Here we discuss a few studies that used different ML algorithms and parameters to measure the $pCO_2$ in the coastal and open oceans around the globe.

Alternate to the statistical model, a self-organized map (SOM), a neural network, successfully mapped the $pCO_2$ in the Atlantic subpolar gyre with latitude, longitude, and SST.[28] SOM also predicted the remaining data with better accuracy than linear regression, and an average of 0.15 Gt-C yr$^{-1}$ sink was estimated from 1995−1997.[28] Later, with SST and Chl-a SOM mapped a basin-wide $pCO_2$ in the Northern Atlantic (RMSE of 19.0 $\mu$atm, a perfect speculative interpolation) with gaps in remote sensing data and performed better when climatological SST and Chl-a were filled in the gaps.[29] Due to the large gap in remote sensing data, Friedrich and Oschlies[30] mapped the $pCO_2$ in the Gulf of Mexico (basin-wide) using voluntary observing ship (VOS) observation (VOS data contains ∼740,000 line measurement of SSS, SST and $pCO_2$ collected in the region 10°S to 70°N) and Agro float data (SST and Chl-a). Notably, the use of Agro data reduced the RMSE in predicting the annual cycle $pCO_2$ by 42% (RMSE of 15.9 $\mu$atm). Further the accuracy of estimating $pCO_2$ can be increased with more Agro floats evenly distributed in the region. Likewise, in the same Northern Atlantic basin, the SOM mapped the $pCO_2$ and constructed the nonlinear relationships between marine $pCO_2$ and three biogeochemical parameters.[31] Satellite-derived Chl-a and SST (NCEP/NCAR) along with MLD and measured $pCO_2$ at a range of 208 to 437 $\mu$atm (collected during 2004 to 2006) the SOM had RMSE of 11.6 $\mu$atm in estimating $pCO_2$ similar to the in situ measurements.[31] Thus, use of SOM proved to be better in estimating $pCO_2$ and had an advantage over other models by avoiding segregation of regions into basins to drive the relationship between the variables and useful in measuring $pCO_2$ in large regions. Similarly, using simple empirical relationship among the carbonate chemistry and remote sensing (SST, Chl-a, and wind stress) data, SOM estimated the $pCO_2$ for the North American Pacific and characterized the 13 biogeochemical subregions, with estimated <20 $\mu$atm root mean squared deviation of $pCO_2$. Also, the model suggested the carbon sink in these regions over a period of 1997−2005 was about ∼14 Tg C yr$^{-1}$ based on the estimated $pCO_2$ valve and wind speed (satellite data).[32] Mapping $pCO_2$ on a global scale remains challenging, as the model created for one region will not perform well for other regions. Moreover, the addition of SSS in the training data improved the performance of SOM in estimating $pCO_2$ in the North Pacific Ocean. The SOM estimated values were paired with the in situ measurement and accurately reproduced the $pCO_2$ values in several time-series locations. Similarly, monthly $pCO_2$ estimation by SOM was similar to the Lamont−Doherty Earth Observatory measurements.[33]

Likewise, a feed forward neural network (FFNN) estimated that the $pCO_2$ reasonably agreed with the in situ measurement with the same parameters SST, Chl-a, latitude, and longitude. The monthly variations of $pCO_2$ with RMSE of ∼6 $\mu$atm were measured using the MODIS-Chl-a and SST data. Also, NN estimated the offshore and onshore $pCO_2$ (13.0 and 12.05 $\mu$atm, respectively) with some uncertainties associated with MODIS data and NN algorithm.[34] A data set was created using FFNN with all the necessary parameters, which helped to estimate the global carbon budget.[35] The data set was reconstructed from the surface ocean $CO_2$ measure from Atlas version 2.0 including monthly distribution of $pCO_2$ world surface oceans. With a data set similar to that in ref 35, a technical note suggested that the comparative analysis showed SVM a better performance in mapping global surface $pCO_2$ followed by FFNN, which took a long time to train. While SOM was the least, SOM had the advantage of fast prediction by training and relabeling, depending on data scaling, which may cause nonsense predictions.[36] In the case of the tropical Atlantic Ocean with the same type of satellite parameters (SST, Chl-a and SSS), the FFNN model has better prediction accuracy (RMSE of 8.7 $\mu$atm) than the linear regression.[37] Also, the regression tree algorithms showed the satellite driven $pCO_2$ values (based on Chl-a, SST, and dissolved organic matter) correlated ($R^2$ of 0.827 and prediction error 31.7 $\mu$atm $pCO_2$) well with ship-based $pCO_2$ measurement. Moreover, $pCO_2$ predicted with satellite-derived salinity was coherent with shipboard measurements. The regression tree model also determined the seasonal air-sea flux of $CO_2$, which was similar to that of biogeochemical models. The tree also predicted that the regional environmental parameters influence the regional spatial distribution patterns of $pCO_2$.[38]

Moreover, the global ocean $pCO_2$ was estimated by FFNN with climatological and Surface Ocean $CO_2$ Atlas (SOCAT) with predictors like SST, SSS, Chl-a, MLD, and sea surface height, latitude, and longitude. Also, the NN predicted the seasonal and interannual variability in the global ocean, where large regions with poor coverage have more influence in estimating global $CO_2$.[39] The application difficulties of a model created in one region to map the $pCO_2$ of another region were solved by a Random Forest-Based Regression Ensemble (RFRE).[40] Different ML models were trained and tested with a data set consisting of field-measured $pCO_2$ data (16 years by different groups) and MODIS satellite data like SST, SSS, Chl-a, and $K_d$. Among the ML models, an RFRE showed better performance with high accuracy (∼1 km special resolution) and less root-mean-square difference (RMSD) of 9.1 $\mu$atm for $pCO_2$ at a range of 145−550 $\mu$atm in most of the Gulf of Mexico regions, and the uncertainty of SST and SSS was found to be highly sensitive compared to the Chl-a and $K_d$ in estimating $pCO_2$. The robustness of the RFRE approach performed well when compared with that of the locally trained model in estimating the $pCO_2$ in the Gulf of Maine. This indicates that the RFRE may be applied to other regions with adequate in situ data.[40] The Cubist model identified the spatial and temporal distributions of $pCO_2$ in the Gulf of Mexico region, which showed seasonal $CO_2$ flux in the region closely related to the change in environmental parameters. Cubist also performed better than most ML algorithms with an RMSE of 8.42 $\mu$atm, where SST, SSS, and Chl-a act as essential variables in estimating $pCO_2$. Also, the model divides the Gulf of Mexico into six subregions based on the distribution of $pCO_2$.[41] As predictors play a vital role in the estimation of $pCO_2$, the FFNN model was used to select the predictors based on the mean absolute error from the 11 biogeochemical regions derived by the SOM. Where FFNN had high precision (with region-specific predictors) in estimating global monthly $pCO_2$ with satellite data (1° × 1° resolution), also reduced the mean absolute error to 11.32 $\mu$atm and RMSE to 17.99 $\mu$atm.[42]

Other than $pCO_2$, the global distribution of total organic carbon (TOC) in the seafloor sediment was determined by the K-nearest neighbor (KNN) algorithm.[43] Based on the estimated geochemical and geophysical properties the model indicates about 87 ± 43 gigatons (Gt) of organic carbon are stored in the upper 5 cm of the seafloor.

**2.3. Nitrogen ($N_2$) and Other Chemicals.** Fixed nitrogen is a vital nutrient for all life on earth, and minor geographical variations in nitrogen bioavailability cause huge disparities in primary production, ecosystem dynamics, and biogeochemical cycles.[44] The balance between denitrification, primarily in oxygen minimum zones (OMZs), and $N_2$ fixation by diazotrophs, primarily in (sub)tropical gyres, determines the fixed $N_2$ forms in the oceans.[45,46] The recent estimate of worldwide marine $N_2$ fixation ranged from less than 100 Tg N year$^{-1}$ to more than 200 Tg N year$^{-1}$.[47] Trichodesmium and unicellular cyanobacteria group-A (UCYN-A) diazotrophs and noncyanobacterial diazotrophs have been shown to contribute considerably to $N_2$ fixation as per recent studies.[48,49] According to statistical algorithms, surface solar radiation and subsurface minimum oxygen are critical factors in the geographic distribution of marine fixed $N_2$.[50] Also, it does not clearly explain the decrease of nitrogen fixation when the subsurface minimum dissolved oxygen level is higher than ∼ 150 $\mu$M. Multiple linear regression estimated the global integrated nitrogen fixation was at 74 Tg N y$^{-1}$ with error ranging between 51−110 Tg N y$^{-1}$.[50] Similar results were predicted by the RF and SVR model (global N$_2$ fixation at a range of 68−90 Tg N year$^{-1}$) based on environmental and biological factors. Predicting global N$_2$ fixation is still a challenging task. Measuring methods such as physiological investigations, satellite estimates, extended observations in undersampled locations, and advanced ML algorithms will solve this problem.

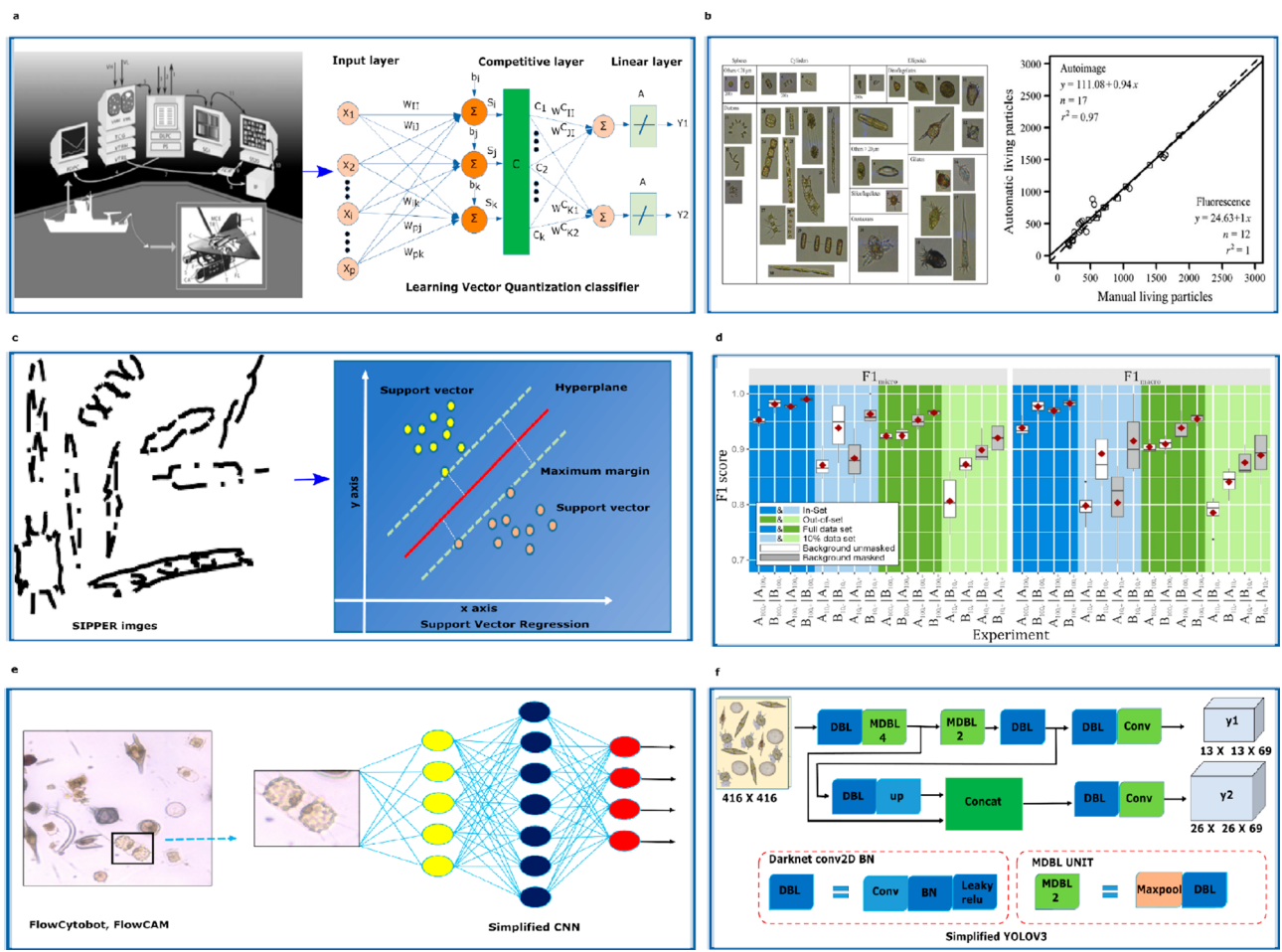

**Figure 1.** Use of ML algorithms to identify, predict, and classify marine phytoplankton from images: (a) An underwater video plankton (VPR) system towed by a ship moved side, up, and down to capture plankton images, and the video recording was analyzed onboard by LVQ (extract image, identify major taxa and its distribution). Reprinted in part with permission from ref 57. Copyright 2004, Inter Research Science. (b) Plankton taxa are grouped in automatic classification with their taxonomic and morphological classes, followed by the comparison manual and automatic classification accuracy for living particles (size range 3−100 μm). The dot and solid line represent the autoimage, and square and dashed line represent fluorescence-triggered samples. Adapted and reprinted in part with permission from ref 61. Copyright 2012, Oxford University Press. (c) Sipper images classified using SVR. (d) Classification performance of VGG16 represented in the boxplot; the red diamonds indicate the mean value, and outliers are indicated by the black dots. Reprinted with permission from ref 72. Copyright 2020, Springer Nature. (e) Use of simple CNN to classify FlowCAM and FlowCytobot images. (f) Plankton classification using YOLOV3.

Likewise, ML addressed the limitation in estimating sedimentary carbonate on the ocean floor and its complex chemistry. The global ocean floor sedimentary carbonate estimation measured the carbonate in the individual site and extrapolated it using an inverse distance weighted technique subjected to a high error rate. To overcome the limitation of data sets, Bradbury and Turchyn[51] used a different ML model to estimate sedimentary carbonate. The model was trained with oceanic physical and chemical properties, including bathymetry, temperature, water depth, distance from shore, tracers of primary production, and data from the global database (ODP/IODP). ML estimated the total amount of sedimentary carbonate formation (1.35 ± 0.5 mol C/yr), which was lower than the previous estimation. Also, ML predicted that 77% of sedimentary carbonate today is mainly driven by anaerobic methane oxidation followed by organoclastic sulfate reduction.

The random forest ensemble model predicted the global ocean surface bromoform and dibromomethane. These halogenated compounds have a short life and affect the ozone in the atmosphere. A data-driven ML algorithm considering the ocean and atmosphere physical parameters along with the biogeochemical factors estimated a global ocean surface emission of 385 and 54 Gg Br per year bromoform and dibromomethane, respectively.[52] Likewise, the sea surface methane disequilibrium (ΔCH$_4$) distribution was mapped by artificial neural networks (ANN) and random regression forest (RRF). Both models successfully predicted local and global spatial patterns, magnitude, and variation of ΔCH$_4$ and estimated the global diffusive CH$_4$ flux of 2−6 Tg CH$_4$ per year from the ocean to the atmosphere. Also, the model showed that the flux was high in near-shore regions where $CH_4$ releases into the atmosphere before oxidation.[53]

## 3. BIOLOGICAL OCEANOGRAPHY

**3.1. Plankton.** Phytoplankton is a vital component of the marine environment, as it plays a crucial role in the biogeochemical cycle. It is a biological criterion for determining the quality of the ocean. Identifying phytoplankton is essential for environmental monitoring, climate change monitoring, and water quality evaluation. Also, understanding the marine plankton ecosystem requires identifying and categorizing them to assess their diversity and abundance. On the other hand, phytoplankton species identification is difficult due to their variability and ambiguity, as there are thousands of micro- and picoplankton species and an imbalance in the distribution of various taxa. Phytoplankton is recognized via imagery and spectrophotometry. Identifying phytoplankton using images is complicated because of the high variation and image quality. Herewith, we discuss a few studies in which machine learning algorithms were used for various tasks, including identification, classification, and database creation.

*3.1.1. Image-Based Classification of Phytoplankton.* Manual analysis of the imagery captured by underwater camera systems is a feasible solution (Figure 1). However, the main difficulties in image classification are image quality, illumination, background noise, angle of the plankton in the image, and deformed objects. Automated image classification using machine learning tools is an alternative to the manual approach. The classification of phytoplankton using the ML started in the early 1990s. The microscopic images were converted into two-dimensional spectral frequency and classified by pattern recognizing algorithm.[54] Later, with preprocessed microscopic images (with Fourier transformation and edge detection), two neural networks and two classical statistical techniques identified 23 dinoflagellates from the images. Among them, a radial basis network outperformed others with 83% accuracy (human 85% accuracy).[55] Moreover, with a combination of Fourier feature with grayscale morphological granulometric, and moment invariants feature, the Learning Vector Quantization classifier (LVQ) classified diatoms and other five planktons with an accuracy of 95%. The LVQ was trained and tested with ∼ 2000 images (six plankton) from the video plankton recorder (VPR).[56] However, the same LVQ with the same features applied to an actual image data set from the VPR system classified *Chaetoceros socialis* with an accuracy of 86% (true positives). Nevertheless, the overall accuracy for seven taxa (e.g., Copepods, Pteropods, Pseudocalanus, Diatoms) were only 60−70%.[57] Also, classification error was high for low abundant taxa. Using a co-occurrence matrix and SVM considerably reduced the error rate for low abundant taxa (20000 plankton image data set that consist seven different plankton categories) more than 50% (especially for *Chaetoceros socialis* where its abundance were low).[58]

Apart from the camera and scanner images, phytoplanktons were identified from the Shadow Image Particle Profiling Evaluation Recorder (SIPPER). The main problem with SIPPER was that many images lacked distinct outlines. With extracted general and domain-specific features, SVM classified diatoms with an accuracy of 79% (with only 64 samples in the training set). The model classified the Trichodesmium with an accuracy of 72.5% in both experiments with 29 and 15 features.[59] A probability value was introduced to evaluate the SVM accuracy, and SVM outperformed (overall accuracy of 75.57%) the C4.5 decision tree and a cascade correlation neural network performance with two different data sets (known plankton images and images with unidentified particles). Also, with a minimal data set, a single SVM outperformed ensembles of decision trees created by bagging and random forests. However, the model struggled to identify unidentified particles in large image data sets.[59] However, the model with the combination of feature selection algorithm (Greedy Feature Flip Algorithm (G-flip)) and SVM identified and measured the abundance of phytoplankton based on the taxonomy from the images taken by custom-built submersible FlowCytobot.[60] A total of 22-category training sets with 131 features were selected from 210 features by G-flip, and SVM was trained with these features, which had an overall accuracy of 88% in classifying independent test sets and 68% to 99% accuracy for individual class categories and also was cross-validated with two-month time-series data from Woods Hole Harbor showing unbiased results concerning manual estimation (random sampling). The model also gave the temporal resolution of phytoplankton abundance and seasonal plankton variability.[60]

The FlowCAM automatic plankton identifier was developed, which uses SVM to classify plankton from images. Even though this automated FlowCAM is an alternate method for manual microscopy, some aspects must be improved to analyze field samples. The classification accuracy of SVM was improved by up to 86% when the images of nonliving objects were eliminated by an automated step.[61] SVM also identified the misestimation of the biovolume of chain-forming diatoms by the current automated method when the biomass of chain-forming diatoms were more than 20% in the sample (estimated using >500 samples). Such classification methods can be used to assign a taxon and simultaneously estimate the biovolume of plankton. Moreover, a minimal difference was observed while comparing the manual estimation.[62] This slight difference was due to the preservation and inaccuracy associated with the automated classification. However, these two approaches had similar results while identifying the seasonal variations in the abundance, biomass, and diversity of plankton in the Cantabrian Sea time series data.[62] Then different features like general and robust were combined by nonlinear Multiple Kernel Learning (MKL), and the use of three kernels (linear, polynomial, and Gaussian kernel functions) showed high recall and precision (90% and 9.91%, respectively) for WHOI data set from Woods Hole Harbor water. Also, it performed better than one kernel and SVM. The only limitation of this model is that it has low efficiency with extremely imbalanced data sets.[63]

Few studies have used neural networks to classify plankton. A Deep Convolutional Neural Network (D-CNN) using rotational and translational symmetry features successfully classified plankton with high accuracy and effectiveness from the PlanktonSet 1.0 image data set (121 classes of plankton raw images captured by the In Situ Ichthyoplankton Imaging System-2 (ISIIS-2)).[64] The two conditions were implemented in CNN layers, i.e., to ensure each convolutional layer can learn complex image patterns, and the receptive field of the top layer should be no greater than the image region. In addition, the inception layer was developed to handle images of various sizes.[65] Most data sets, like the WHOI-Plankton data set, had the class imbalance problem, leading most models to classify only the major classes and neglect the minor class during classification. The transfer learning-based CNN and the class normalization function solved this problem where the class-normalized data set was created by reducing the large-sized classes by random sampling and the CIFAR10 CNN model was trained with the normalized data set. To reflect the actual data size, transfer learning was applied to the CNN trained with normalized data by retraining with the original data set, which improved the class imbalance problem and class population bias in classifying planktons. Also, CNN with normalized data and transfer learning had better overall accuracy than CNN with different data augmentation with transfer learning and CNN without transfer learning.[66]

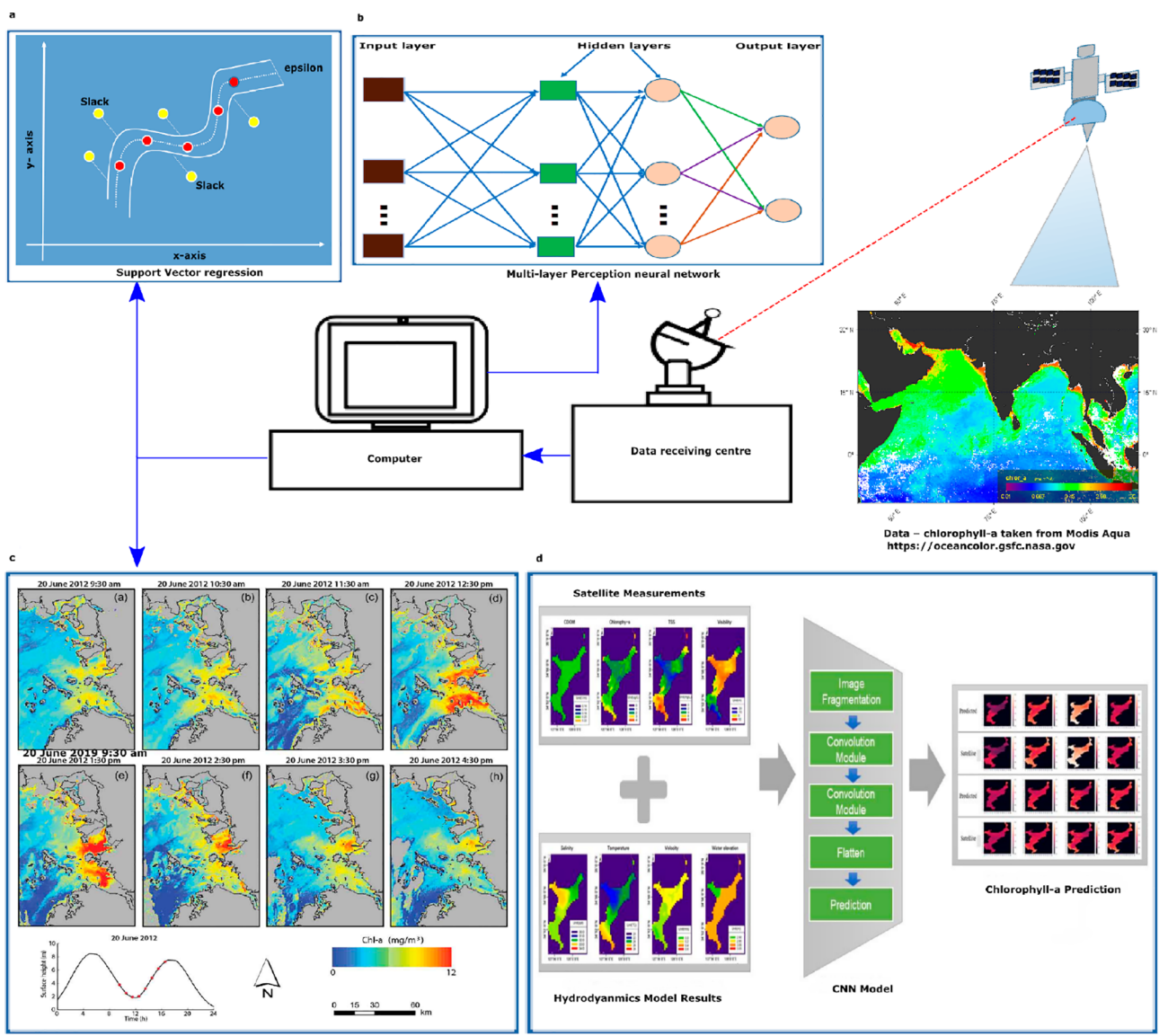

**Figure 2.** Use of ML algorithms to measure and estimate the ocean surface chlorophyll-a from satellite data. (a) Prediction of Chl-a using SVR and (b) multilayer perceptron neural network. (c) SVR model prediction for the hourly spatial distribution of chlorophyll-a concentration on the west coast of South Korea. Reprinted in part with permission from ref 76. Copyright 2014, Taylor & Francis. (d) The overall workflow for the prediction of Chl-a using CNN. Reprinted in part with permission from ref 82. Copyright 2021, MDPI.

Also, CNNs trained with two different plankton images (ISIIS and IFCB) had better feature extraction on the third data set when weight transfer was applied and showed high performance in classifying plankton. This method is an advancement in the CNN that provides better or comparable results.[67] Similarly,[68] applications of CNN-based transfer learning approach to extract the features from the openly available plankton image data set and used SVM to classify planktons, which had good accuracy. This showed the transfer learning approach's effectiveness in coping with the large scale and high variable plankton images data set and the usefulness of the CNN-based feature extraction as an alternative for regular hand-picked features to estimate the commonly available planktons in the water samples.[68] Where the ResNets pretrained with the ImageNet data set were used to obtain features, the model with these deep features better estimated the frequently occurring plankton class in the water sample.[69]

Apart from the extraction and transfer learning, a deep convolutional neural network (DCNN) was used to reconstruct the phase contrast image. A portable flow cytometer with coherent lens-free holographic microscopy captures diffraction patterns of objects that flow through the microfluidic channel at a rate of 1000 mL/h. DL phase-recovery reconstructed the diffraction patterns, and the images were reconstructed in real-time. This device showed high efficiency in capturing images from the ocean samples and showed similar results to the California Department of Public Health in measuring the abundance of toxic plankton *Pseudonitzschia* in six Los Angeles public beaches. This portable imaging flow cytometer may be used for continuous economic and portable monitoring.[70]

Another functional adaptation in CNN was the combination of several fine-tuning CNN models that were trained to different strategies and proved to have better performance than a single CNN in classifying plankton (from 3 plankton and 2 coral data sets). Even though the stand-alone DenseNet was found to be the best model for the target data sets, the ensemble model has considerable improvements. Also, the final proposed ensemble consisting of only 11 classifiers (the number of classifiers was also reduced using feature selection) performed better.[71] CNN also identified the taxonomy of diverse morphological diatoms from the image data sets assembled from two Southern Ocean expeditions. The CNN performance was checked with background masking, data set size and possible changes in image classification performance. The old CNN architect model, VGG16, had better performance and generalizing ability from the given image data set, which was further improved after background masking. Also, CNN showed high performance when the top layer of CNN architect was alone trained extensively.[72]

Likewise, Li et al.[73] introduced a new phytoplankton microscopic image data set (PMID2019), which contains 10819 phytoplankton microscopic images of 24 different classes. The data set was created using the dead cells images, and cycle-consistent adversarial networks (cycle-GAN) were utilized to generate the corresponding living phytoplankton cell images. Also, a few live cell images (only 217 images, including 10 different categories) were included in the data set. The database was tested with different ML algorithms; among them, Fast R-CNN has better accuracy for predicting the location and class of plankton in the images. The PMID2019 database was used to assess ML algorithms' performance that detects plankton from the image.

**3.1.2. Estimation of Ocean Chlorophyll-a from Satellite Data.** Microalgae, especially phytoplanktons, dominate the upper sunlight zone of the ocean, act as the primary source of the oceanic food web, and fix carbon via photosynthesis.[74,75] This phytoplankton has a pigment called chlorophyll-a (Chl-a), which involves photosynthesis and measuring Chl-a as an essential parameter in assessing water quality. Along with Chl-a, suspended particles and colored dissolved organic matter (CDOM) in the surface water can be measured using remote sensing. The recent development of sensors (moderate resolution spectroradiometer (MODIS), sea-viewing wide field-of-view sensor statistics (SeaWiFS), and medium resolution imaging spectrometer sensors (MERIS) and different retrieval algorithms were used to estimate the Chl-a concentration from remote sensing data. However, the accurate measurement of Chl-a from satellite data is still challenging. Here, we have discussed a few studies implementing ML algorithms to estimate the Chl-a from remote sensing data

The accurate estimation of Chl-a and monitoring of the coastal environment are still challenging even with three decades of satellite observation. The coastal water quality is influenced by many factors, such as inputs from inland and coastal circulation. The simple numerical methods could not accurately estimate the water quality because different factors, such as suspended particulate matter (SPMs) and CDOM, affect the spectral response. However, SVR algorithms overcome these limitations and estimate the Chl-a and SPMs concentrations in the surface waters of the west coast of South Korea.[76] SVR has shown better prediction accuracy than RF and Cubist, with $R^2$ values of 0.91 and 0.98 for Chl-a and SPMs, respectively. Geostationary Ocean Color Imager (GOCI) satellite data and the field measurements data (as reference) were used for training and testing. Where SVR showed the ratio of band 2 to band 4, bands 6 and 5 were the critical variables in predicting the Chl-a and SPMs concentrations when GOCI-derived radiance data were used (Figure 2).

Similarly, SVR successfully estimated the surface global ocean Chl-a concentration using the NASA bio-Optical Marine Algorithm Data set (NOMAD) as a training data set.[77] SVR reduces the image noise and improves the cross-sensor consistency; it also produces consistent results with different sensors (SeaWiFS, MODISA, and MERIS) and performs better than the band-ratio OCx approaches (evaluation with various sensor data) even though the SVR performance was statistically slightly less than the empirical color index (CI) algorithms for Chl <0.25 mg m$^{-3}$.[78] The SVR model showed extended applicability to international waters, from the CIs 0.01−0.25 mg m$^{-3}$ (about 75% of the global oceans) to 0.01− 1 mg m$^{-3}$ (96% of the global ocean). Also, compared to the NASA hybrid Ocean algorithm (OCI), SVR was simple and avoided the complexity of mixing two algorithms, thus shown as a possible alternative method for global chl-a estimation.[77]

However, Extra tree, a deep learning model, successfully measured the Chl-a concentration from the sea surface reflectance data over West Africa.[79] The ESA Ocean Color Climate Change Initiative satellite sensor data was used as a training data set, whereas the MODIS sensor data set was used to validate the model. The Extra tree shows high accuracy (96.46%) and low mean absolute error (0.07 mg/m-3), and the model performed well with mixed or single sensor data. Also, the estimated Chl-a values by the Extra tree model were consistent with upwelling phenomena observed in this area.[79] However, using Bayesian maximum entropy (BME) and SVR improved Chl-a estimation from satellite reflectance data by reducing the non-negligible uncertainties.[80] In the initial model building step, SVR performed well with higher accuracy than other ML algorithms in estimating the Chl-a concentration from MODIS Remote sensing reflectance ($Rrs$) at 412, 443, 488, 531, and 678 nm data with R$^2$ values varied between 0.708 to 0.907 for training and validation, respectively. Then this SVR estimated Chl-a concentration was processed using BME with the incorporation of inherent spatiotemporal dependency of physical Chl-a distribution, reducing 56% of the mean non-negligible uncertainties. The BME/SVR also estimated the daily mean Chl-a concentration, which varied between 1.663 to 3.343 mg/m$^3$.[80]

Besides the SVR and deep learning models, NN was also used to estimate the Chl-a from satellite reflectance data. Among the tested ML algorithms, ANN performed better in estimating the Chl-a, suspended solids, and turbidity from the Landsat reflectance data. Moreover, compared to standard Case-2 Regional/Coast Color" (C2RCC), ANN had high accuracy in estimating Chl-a from satellite (91% accuracy with a low RMSE value of 2.7 $\mu$g/L) as well as from the *in situ* reflectance data sets (89%).[81] CNN has also been used to estimate the spatial and temporal distributions of Chl-a in the Korean bay.[82] Two CNN models were built (which use different dimensions of satellite images), trained, and tested with satellite color images data (Chl-a, total suspended sediment, visibility, and CDOM) and hydrodynamic data (water level, currents, temperature, and salinity) generated from the hydrodynamic model. CNN-II with a 300 times large data set (7 × 7 segmented image) showed better prediction accuracy with $R^2$ exceeding 0.91 and low average RMSE (0.191). Also, the model predicted that CDOM plays a vital role in estimating the spatial-temporal distribution of Chl-a from the satellite color data. Likewise, a neural network model named Ocean Color Net (OCN)) with match-up data sets showed to improve the Chl-a estimation on the surface and within the productive zone of the Barents Sea using satellite imagery data.[83] A new spatial window-based match-up data set was created by matching depth-integrated in situ Chl-a concentration with the multispectral remote sensing images from Sentinel-2. After the removal of the erroneous samples in the match-ups data sets based on satellite reflectance, the OCN was trained and tested with the match-ups data set. OCN outperforms the existing ML models (Gaussian Process Regression (GPR), Ocean Color (OC3) algorithm, Case-2 Regional Coast Color (C2RCC), and the spectral band ratios) with less mean absolute error. Also, the spatial window and depth-integrated match-up data set improved the performance of the OCN by 57%. This model showed the ability to produce a realistic chl-a map by capturing the fine details and being able to observe the small change in distribution.[84]

*3.1.3. Numerical Data Set-Based Plankton Classification.* ML algorithms like M5P and regression trees integrated with the WEKA was used to study phytoplankton dynamics in station RV001 in front of Rovinj, open Northern Adriatic Sea (NAS).[84] The first model (M5P) identifies the factors that influence the phytoplankton abundance in the NAS from the 28 years (1979−2007) data set containing phytoplankton concentration along with the physicochemical parameters (salinity, temperature, river flow (Po River), month, year). The M5P model predicted salinity and temperature as essential factors that influence the phytoplankton abundance, and a significant change in phytoplankton dynamics occurred at the NAS in 1998 and three years (1985, 1989, and 1993) before 1998 (coefficient correlation of 0.7) where the second model successfully forecasts the phytoplankton concentration with a good coefficient correlation of 0.82. This type of model may be helpful in predicting phytoplankton concentrations with nutrient information. However, a consensus model (weighted average prediction error (WA-PE) model) created by combining four single-model predictions (SVM, RF, Boosting, and generalized linear models) successfully predicted phytoplankton species with low classification error from the phytoplankton (eight) presence and absence data.[85] The model WA-PE showed a low classification error in classifying *Akashiwo sanguinea* and *Dinophysis acuminata* (10% and 38%, respectively).

Apart from the environmental variables influencing the phytoplankton variability in the time-series study, the uncertainty caused in the laboratory was also predicted by two ML algorithms (Bray−Curtis distance and pairwise permutational multivariate analysis of variance).[86] The Bray−Curtis distance showed that in long time-series phytoplankton variability studies significant variation was caused by different experts handling the sample and fixatives used. Also, PERMANOVA showed significant variations observed between the type of preserving agent (glutaraldehyde and Lugol's solution) and between the taxonomists involved in the study.[87]

Genetic programming (GP) efficiently identified the strong association between a rise in water temperature with reduced net primary productivity (NPP) in the oligotrophic ocean.[87] The 27 year Bermuda Atlantic Time-series Study (BATS) data set contains NPP and environmental parameters. GP showed reduced NPP due to warming and weakening of vertical mixing in the upper water column, which reduces nutrient availability (light and nitrogen). This model indicates the necessity to have a long-term monitoring study with advanced omics techniques in the oligotrophic region of the ocean to better understand the early trends and predict future oceanic conditions. A neural-network-derived quantitative niche model predicted Pico-phytoplankton lineages were separated into latitudinal niches based on the cell size and showed increased biomass along the temperature gradient in low-latitude regions.[88] The model also predicted (based on the global data set) a high concentration of cells found in the North Atlantic (above 45°N), around the North Pacific Current, and a band near the southern subtropical convergence zone. In comparison, oligotrophic tires and polar regions showed a low concentration of cells. The model also predicts future increases in seawater temperature in low-latitude regions may lead to an increase in the biomass of picophytoplankton, which is also supported by the elevated upper-ocean nutrient recycling and lower nutrient requirements of phytoplanktons.[89]

*3.1.4. Harmful Algal Bloom (HAB).* HAB is the rapid proliferation of microscopic algae or phytoplankton (including blue-green algae) and accumulates toxic or other noxious substances. Some HAB produces nontoxic compounds that react with reactive oxygen species, polyunsaturated fatty acids, and mucilage. These HAB are lethal to fishes and cause faunal mortality via high biomass accumulation, leading to oxygen depletion.[89] HAB causes fish deaths worldwide, and these issues appear to be frequently growing. Developing early warning systems is one approach to reducing their effects on people's health and livelihoods.[90] Fisher's linear discriminant analysis (LDA) classified algal blooms based on spectral properties at the order level. The spectral properties of 53 different unialgal cultures were used as training data. LDA spectral properties with a leave-one-out cross-validation method and cross-examined with mixed algal culture, LDA excellently classified cyanobacteria from other algal groups, and the accuracy ranged between 81.5% and 100% for each algal group. Also, LDA had a low error rate of 9.3% or no error rate in identifying dinoflagellates and cyanobacteria, respectively. LDA had high accuracy in identifying dinoflagellates (90.7%), cyanobacteria (100%), and other algal groups (96.3%).[91] Likewise, a researcher[90] developed two early warning systems to detect harmful algal bloom using RF. A three-year field sensor data set of temperature, salinity, DO, pH, Chl, shellfish ban, and fish kill occurrences from Bolinao-Anda, Philippines, were used to train the RF model. The RF model had an accuracy of 96.1% for the fish kill with a decrease in DO, higher temperature, and salinity as essential factors. The model had 97.8% accuracy in detecting shellfish ban, influenced by a decrease of DO, low salinity, and higher Chl conditions. These models might have applications in the real-time monitoring of HABs in marine environments.

**3.2. Zooplankton Classification.** Zooplankton (ZP) are highly abundant and play an important role in the ocean's biogeochemical cycle. These ZP are highly diverse and range from microzooplankton to metazoans. Classification of ZP through image is much more challenging than that of phytoplankton, as ZP have different sizes and morphology. Even though they have distinguishable differences between the groups like copepods and euphausiids, they possess remarkable similarities between the closely related genera (e.g., *Calanus* spp. and *Paracalanus* spp.), which makes the identification of ZP highly challenging. Also, manual identification requires a skilled taxonomist and is a time-consuming process. So automatic identification through images is an alternate methodology. The identification and measuring of the size of ZP through images started in the 1980s from Silhouette imaging[92] and images.[93] Early ML algorithms like discriminant analysis were used to identify the frequently observed eight ZP taxonomic groups from images (from the coastal waters of England, which had an accuracy of 90%)[94] and flow-through sampler images.[95] All the above methods have disadvantages like the image quality (orientation of objects and low contrast) and small data set.

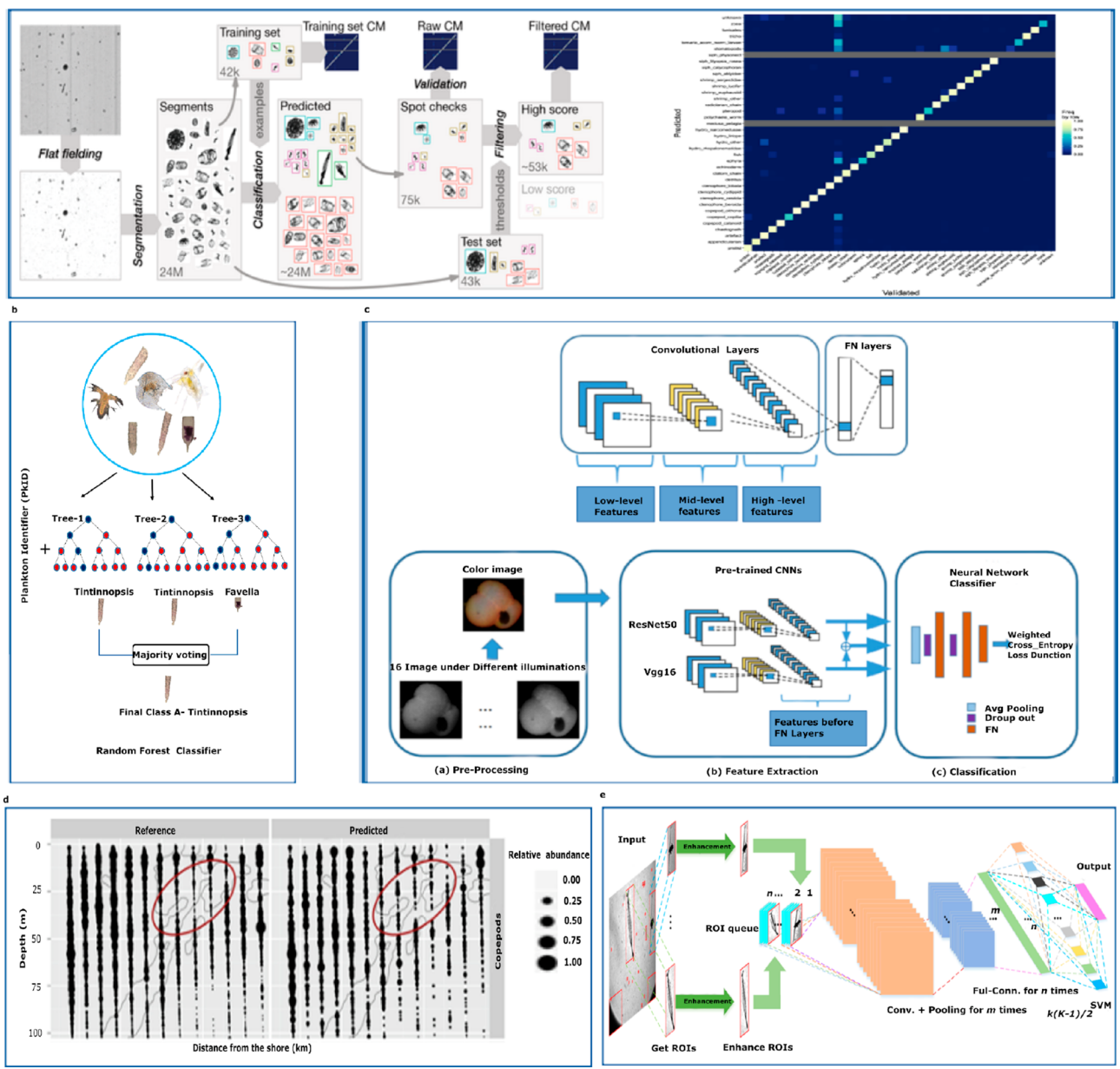

**Figure 3.** ML algorithms are used to identify or classify zooplankton through images captured by various underwater devices. (a) Complete workflow for identification of ZP, starting with image segregation, training, and classification by CNN, followed by validation (classification accuracy), which distinguishes the samples into low and high probability and a confusion matrix showing the model classified images into 108 classes from 75,000 random ZP images. Reprinted in part with permission from ref 111. Copyright 2018, ASLO. (b) RF used for classifying tintinnopsis. (c) Basic CNN and pipeline for classification of foraminifera. Reprinted in part with permission from ref 116. Copyright 2019, Elsevier. (d) Automatic prediction of the spatial distribution of copepods well correlated with reference distribution, x-axis distance from the shore, and the y-axis represents the depth in meters. Reprinted in part with permission from ref 108. Copyright 2016, Elsevier. (e) Pipeline representing the steps and ML algorithms used for automated plankton identification and counting. Reprinted in part with permission from ref 113. Copyright 2019, PLoS.

NN algorithms such as a backward error-propagation neural network identified ZP with a reasonably small data set[96,97] and learning vector quantization classifier (LVQ) with Fourier feature classified the images (8000 images) captured by the video plankton recorder (VPR) (Figure 3).[98] Also, a combination of different neural networks was used for pattern recognition.[56] LVQ with extracted ROIs and different neuron numbers measured the size and abundance of ZP from large image data sets.[57] Even with the neural networks, the accuracy was less in estimating abundance when the taxon relative abundance is low, which was considerably improved (50%) by the SVM classifier with the co-occurrence matrices as a feature. Also, the model showed a reduced error rate with a data set consisting of 7 categories of manually sorted 20000 plankton images captured by VPR.[58] Similarly, a new discriminant vector forest algorithm which is a combination of LDA, LVQ, and RF, identified 2000 items per second at an accuracy of 75% from the ZooScan images,[99] and ZooScan captures ZP images with a 2400-dpi resolution and the model was chosen based on the validation.[99] Along with SVM, decision trees and other unsupervised algorithms are also used to identify ZP, which have 70−80% accuracy.[100]

Although these methods showed considerable accuracy, they had some disadvantages that required manual sorting of images, trained only with lab-preserved samples, poor image quality, and slow computation power. To overcome this, a fully automatic dual classification system was utilized[101] where the planktons were identified by an LVQ using shapes as features followed by SVM using texture-based features. The estimation of abundance based on the dual classification was close to manual results.[101] In 2007, ZooImage was created to predict the taxonomy of preserved ZP[102] is a unique integrated system used to import, segment images, extract features, train, and classify the data (ZP).[103] This ZooImage (automated system) with RF showed difficulties in classifying field samples, where the accuracy dropped to 63.3% from 81.7% when the other nonliving substance was removed from the samples. Also, the model predicted ZP size is an essential feature in classification.[104] Later the combination of ZooScan, Zooprocess, and Plankton Identifier software (PkID) with RF (to identify) and manual validation identified ZP with better accuracy. RF was chosen on the basis of the performance compared with six other classifiers. The PkID had a better performance with RF and classified ZP moderately with 80% accuracy from the Villefranche time-series data sets (Validation), and the performance was slightly improved second iteration.[105] Therefore, the model was not amply used in ecological studies.

All these models predict only the highly abundant taxa, and estimating of low abundant rare taxa diversity and composition is still challenging. A semiautomatic model with a naïve Bayesian classifier (NBC) and manual reclassification overcomes these difficulties. NBC presents the images with low predictive confidence to manual reclassification, which improves the accuracy in both unbalanced and balanced training data sets. The NBC predicted rare taxa at high accuracy from the 154289 zooplankton images (East China sea), which helps estimate diversity indices and ecological studies.[106] Moreover, with simple geometric features, the SVM model had better performance in classifying ZP.[107] Moreover, ML algorithms used to solve ecological conclusions (distribution patterns). The ecological conclusions obtained using Zooprocess and PkID post-processed data set were tested with manually sorted fully automatic data set. The distribution predicted by the RF model was similar to the reference distribution. With the RF model defined probability score, the accuracy was increased by 16% after removing the class with a probability threshold of 1% error rate (84% accuracy). Most importantly, the model automatically predicted the difference in the distribution of abundance over a larger region and the pattern between day and night.[108]

Earlier, deep learning methods were used to detect and classify ZP, but a large data set is required for Deep Learning, CNN, and ensemble models. The ZP image data set was relatively small compared to previous studies and consisted of a class imbalance problem (lack of images for low abundant ZP). Where the low abundant taxa were not predicted by most of the ML algorithms, these problems were addressed using a deep learning architecture, "ZooplanktonNet", an automatic classifier that reduces the overfitting caused by a lack of data by applying data augmentation. Using general and representative features rather than predefined extraction algorithms, CNN classified ZP with 93.7% accuracy.[109] Also, a deep residual network classifies plankton from images with an accuracy of 95.8% (at 9.1 ftps).[110] Later, CNN (a spatially sparse) identified ZP with an accuracy of 84% (recall rate of 40%) from 2.4 million images captured by the advanced imaging system *In Situ* Ichthyoplankton Imaging System (ISIIS). CNN identified 108 plankton from a 40 h underwater image data collected from the eight transects in the northern Gulf of Mexico. Also, the accuracy was increased to >90% when rare taxa were removed.[111] Similarly, the YOLO V2 model performed considerably well (with a precision of 94% and a recall rate of 88%) in classifying ZP from the holographic images with a sharpness assessment score equal to 0.6 or more.[112] This approach demonstrated that the efficacy of CNN could be applied to various plankton and biological imaging classification systems with eventual application in ecological and fisheries management. In combination with SVM with different CNN, models showed increased classification and recall accuracy (7.13% and 6.41%, respectively). Where CNN extracted the ROIs (from the plankton images), the ROIs were enhanced by removing background noise, and SVM was used for classification. Among CNN, ResNet50 with multiclass SVM showed the best accuracy and recall (94.52% and 94.13%, respectively).[113] This model provides information about the selection of algorithms.

However, CNN had difficulties identifying ZP when the images were rotated at a certain angle. This rotational variance was overcome by the combined model (CNN and SVM), which had a mechanism that mimicked human eye movement to extract features (Cartesian coordinates and polar coordinates). Then these vectors were fused and used to train the convolutional learning, later classified by SVM. The model (DenseNet201+ polar + SVM) had high accuracy (94.91%) and recall rate (94.76%) (validated) against the CIFAR-10 image data set.[114] Moreover, a sparse CNN identified 64 ZP taxa from the ISIIS collected on the Oregon coast, with an accuracy and recall rate of 83% and 56%, respectively. The data set used for training consists of 52 million images of plankton like copepods, protists, and gelatinous organisms.[115] Similarly, CNN identified six planktic foraminifera with better precision and recall accuracy (80%). The CNN trained with a database containing light microscopic images of six pale-oceanographic important planktic foraminifera (at 16 different illumination angles) had better performance than human beginners and experts. This automatic identification may lead to the development of an automatic robotic system to identify foraminifera and reduce time, which allows human experts to focus on the morphotypes or intergrades. Also, this method aids humans in gaining more profound knowledge of foraminifera taxonomy in less time.[116]

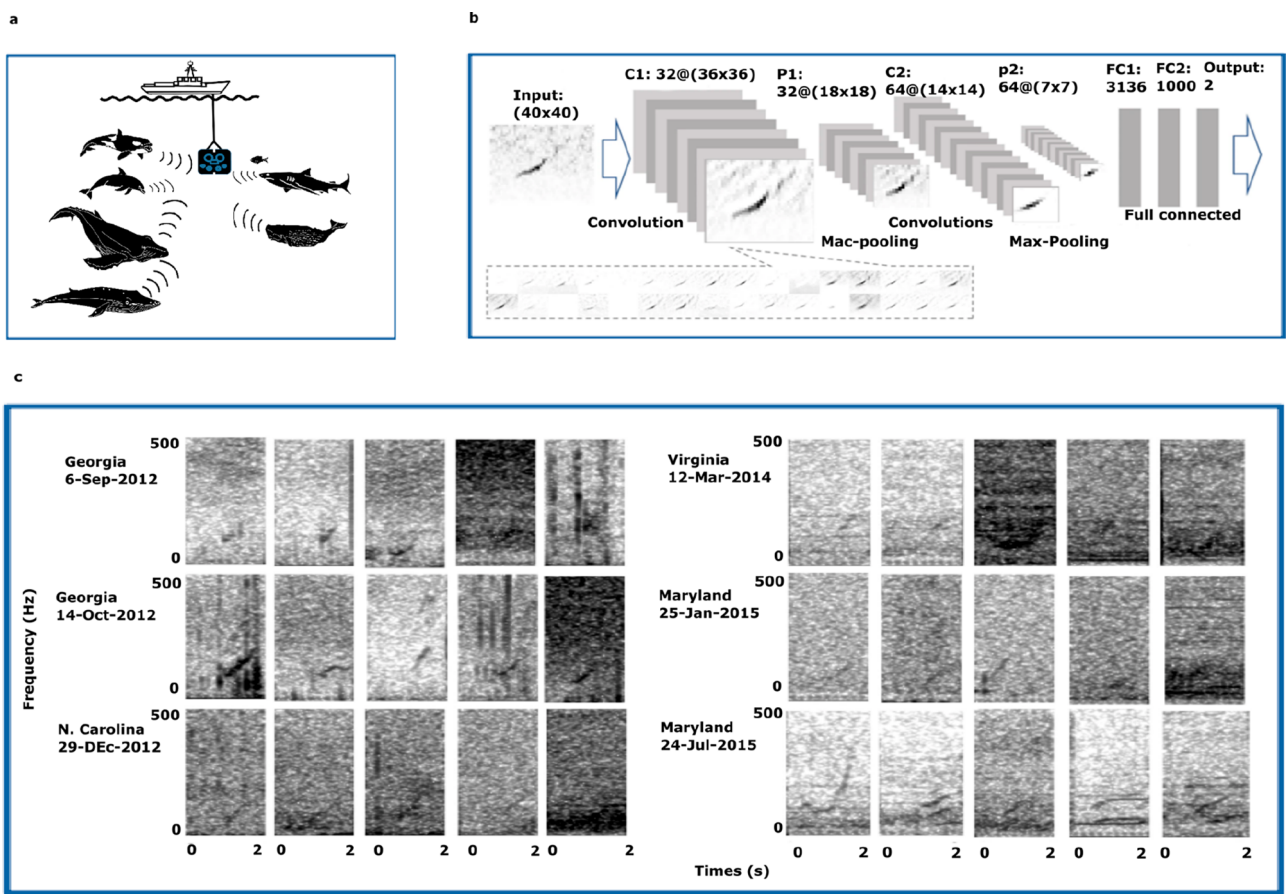

**Figure 4.** ML algorithms are used to identify and understand the behavior of marine fishes and mammals from acoustic data, (a) Graphical representation of acoustic data recording. (b) Workflow to detect right whale calls using LeNet CNN where feature maps were generated by convolution and max-pooling and (c) prediction of right whale upcalls by deep net with probability >0.8. Reprinted in part with permission from ref 131. Copyright 2020, Springer Nature.

The main idea behind the ML algorithms is to apply them in the field and to measure them in real time. A novel mobile robotic tool, "AILARON" (Automatic Underwater Vehicle) was created to characterize the upper water column biota. The images captured by the silhouette camera were classified by deep learning and grouped on the basis of the probability score. This processing pipeline consists of imaging, supervised machine learning, hydrodynamics, and AI planning, which process each image in an average of 3.852 s. AILARON may be useful to enhance the knowledge of plankton communities and their Spatiotemporal distribution patterns and have great importance in ecosystem surveillance and monitoring global change.[117]

Apart from the identification and estimation of ZP, ML algorithms were used to predict the changes in ZP abundance and occurrence based on environmental conditions. The Boosted Regression Tree (BRT) model predicted the space and time measurements of six zooplankton abundance in SO (Copepods, Foraminifera, *Fritllaria* spp., *Oithona similis*, and Pteropods). Based on the abundance and environmental data sets, the model predicted that over two decades (1997−2018) the environmental conditions changed in favor of copepods,

Foraminifera, and *Fritllaria* spp. (increased the abundance by 0.72% per year). Also, the conditions in the Ross Sea shelf regions have significantly deteriorated the pteropods' abundance.[118] Moreover, RF with manual correction of ZooImage successfully classified all major ZP classes with precision and recall ranging between 0.07% and 0.20% and 0.82% and 0.94%, respectively. Based on the RF and manual correction, 25% of the total annual abundance in the Mediterranean Sea was recorded in April alone and influenced by wind gusts, nitrate availability, and water temperature, while the concentration of chl-a and ZP was ambiguous. Moreover, the rise in seawater temperature was in sync with the low ZP annual abundance after 2010.[119]

The high detection and classification of ZP can be approached, viz., a high quantity of features and optimization of classification models to prevent feature loss. The detection of rare ZP taxa through deep learning has some limitations. One such limitation is the class imbalance and reduction of plankton feature loss in neural networks. Eight different rare ZP were identified using NBC (using posterior probability and predictive confidence value), with accuracy ranging between 0.18 and 0.87.[106] Many studies used data augmentation to create training data sets by capturing images in different brightness, image orientation angle, etc.[109,112,114] One such data augmentation technique is the use of a Cycle-consistent Adversarial Network (CycleGAN). The data generated by CycleGAN was successfully classified by a densely connected

YOLO V3, which outperformed previous state-of-the-art models with mean Average Precision (mAP) of 97.21% and 97.14% (two experimental data sets with varied numbers of rare taxa). The model also improved the accuracy of detecting rare taxa by an average of 4.02% and has the potential to be included in autonomous underwater vehicles for real-time identification and plankton ecosystem observation.[120] The model proposed by Gorsky et al.[105] ZooScan with ZooProcess was successfully implemented in the identification and estimation of ZP abundance in the bay of Villefranche-sur-Mer, France.[121] This shows that ML algorithms are a potential tool not only for the identification of ZP but also used to solving ecological problems. Still, few improvements are required in this field to achieve full automation to detect rare taxa and morphologically similar species.

**3.3. Identification and Classification of Fishes and Mammals Using Acoustic Data.** Most marine mammals and fish produce acoustic (sounds) to communicate within a group or species or to locate prey. This acoustic has a different frequency range, and the frequency differs based on the animal that produces it. As an alternate method to visual identification, these acoustic data were used to study the animals' behavior. Automatic analysis of acoustic data and identification of individual clicks produced by marine mammals are challenging. The acoustic signals are influenced by many factors like the depth in which the animal dwells, orientation and the distance from the hydrophone.[122,123] Few studies used wavelet transformation (mathematical models) to analyze clicks.[124−126] This method faced difficulties in characterizing dive clicks. The use of ML in the identification of animals from their acoustic data started in the early 90s. A back-propagation neural network (ANN) accurately discriminates acoustic from the individual orca as well as the same calls from other whales within the group.[127] It also discriminates the *Orcinus orca* behavior based on the acoustic data. This approach not only enhances knowledge of whale behavior based on acoustic calls without the need for visual confirmation but also reduces the time that inexperienced observers take to identify the whales (Figure 4).

Similarly, the Bienstock, Cooper, and Munro (BCM) unsupervised neural network successfully classified different mammal sounds, even those recorded from different geographical regions.[128] Likewise, the unsupervised ML-NN (a self-organizing network) detects and categorizes the vocalization of false killer whales without any predefined categories.[129] The 2D data set contains short measures of duty calls, and the peak frequency of false killer whale vocalizations was analyzed using two-NN, where the competitive learning (first neural network) recognizes vectors that are frequently presented in input vocalization and categorize them in class patterns. The Kohonen feature map (second network) provides pattern relationships (graphical representation) for the outputs defined by the first NN. The model performed well in categorizing vocalization and the ability to classify the vocalization of other mammals.[130] Also, a radial basis function network model (a two-layer neural network) successfully separated the individual whales' clicks from a group of hunting sperm whales' recordings.[126] The NN trained with six individual male diving clicks, consisting of five short and one complete diving click. A wave-based local discriminant basis extracted the features (clicks), and the extracted features were used to train the model with 50 clicks from each data set, and the rest of the clicks were used for testing. The model classified the short and diving clicks with 90% and 78% accuracy, respectively.[127]

Similarly, the bioacoustics behavior of *Physeter macrocephalus* (sperm whale) was effectively classified using CNN based click detector.[130] Based on the presence and absence of clicks in the spectrogram, CNN successfully classified 650 spectrograms with an accuracy of 99.5%. Furthermore, a trained CNN-based click detector successfully classified three types of task, i.e., coda type classification, vocal clan, and classification of individual whales using Long-term memory and gated recurrent unit recurrent neural networks. The CNN showed high accuracy in classifying 23 and 43 coda types from the Dominica data set (8719 codas) with an accuracy of 97.5% and the Eastern Tropical Pacific (ETP) data set (16,995 codas) with an accuracy of 93.6%, respectively. Also, the model has an accuracy of 95.3% for the Dominica data set (for two clans) and 93.1% for the ETP data set (for four types of clans) in classifying the vocal clan. Moreover, the model has 99.4% accuracy in identifying individual whales' clicks. These results demonstrate the feasibility of applying CNN to classify sperm whale bioacoustics and learning fine details of whale vocalisations. In advance, deep neural networks successfully detect the vocalizations of the endangered North Atlantic right whale *Eubalaena glacialis*.[131] Different deep learning models were tested, where the LeNet performed better with the lowest false positive rate. Three data sets from the Detection Classification, Localization, and Density Estimation of Marine Mammals (DCLDE 2013 consists of right whales acoustic recordings from the coast of Massachusetts in 2000, 2008, and 2009 recorded by the NOAA NorthEast Fisheries Science Center and the Cornell Bioacoustics Research program. The original data was recorded with six or ten devices, where for the workshop only a single channel was converted and used, which consist of 7 days of right whales' upsweep and gunshot calls) data set, MARU deployments data set, and keggle (whale competition-Massachusetts contains recordings of upwelling calls of right whales recorded by the 10 autodetection buoys implemented in the Massachusetts Bay. The autobuoy has a frequency range of 15−585 Hz which was similar to the right whale upcalls) used for the study. LeNet not only had significant high precision and recall but also had lower false positives than the algorithms presented at the DCLDE 2013. CNN trained with recordings from one geographic location over a period of time, was able to recognize calls spanning many years and across the species' range with a low percentage of false-positive rate. It is also simple to integrate into current software, allowing researchers to learn more about threatened species.

ML algorithms other than neural networks were used to study the acoustic data. The clicks produced by either one or more individuals of the following species, i.e., Blainville's beaked whales, short-finned pilot whales, and Risso's dolphins, were differentiated by Gaussian mixture models (GMMs) and SVM.[132] The Teager energy operator locates the individual click from the echolocation click recorder, and cepstral analysis constructs the feature vectors for these clicks. Two detectors based on GMMs and SVM trained with the cepstral feature conform or reject the species based on the clicks, GMMs model the time series of independent characters of species feature distribution, and the SVM model differentiates the one species to another by creating boundaries between the species feature distribution. Both models detect the clicks with a lower error rate.

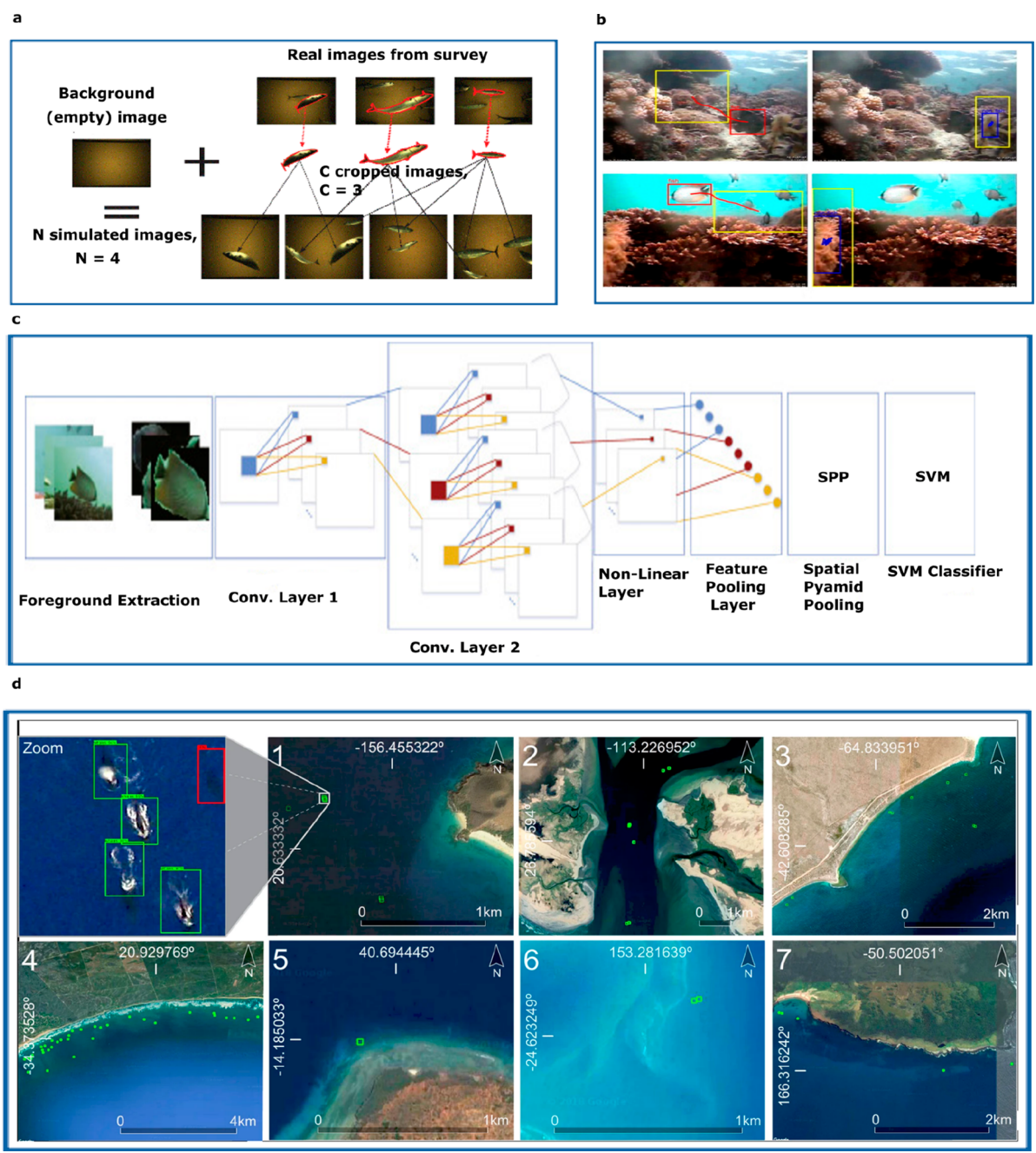

**Figure 5.** Different ML algorithms were used to identify marine fishes and mammals. (a) Images were created that resembles Deep vision photography, where different numbers of fish images were cropped and pasted into the empty background (at a random spot, orientation, and size) and used to train the ML model for real-time identification of fishes. Reprinted in part with permission from ref 141. Copyright 2019, Oxford University Press. (b) Classification of swimming fish from the background and drifting particles, where ML identified fishes (red box) and nonliving particles (blue box) successfully. Reprinted in part with permission from ref 137. Copyright 2014, Elsevier. (c) Workflow proposed for identifying fishes, where CNN extracts features, than pooling and finally classification by linear SVM. Reprinted in part with permission from ref 138. Copyright 2016, Elsevier. (d) Counting of whales by CNN-step-2, based on CNN step-1, which locates the whale in the grid cell green boxes, whereas the red box represents the false negative result. Map data were obtained from Google and DigitalGlobe. Reprinted in part with permission from ref 145. Copyright 2019, Springer Nature.

The sensitivity level between the different abiotic variables and the acoustic density of fishery resources (two different layers) in the Northern South China Sea was estimated using extreme gradient boosting (XGBoost) and RF.[133] The fish density acoustic data from the surface mixed layer and bottom cold-water layer, along with the abiotic variables (temperature, salinity, water depth, nitrite, nitrate, ammonia, and phosphate) were used to train and test ML models. Nautical Area Scattering Coefficient characterized the acoustic data values (NASC), XGBoost predicted the surface temperature, and nitrate had a high sensitivity value with the NASC value at the surface mixed layer (10 m), whereas RF predicted the surface temperature as a significant factor for the surface NASC value. In the bottom NASC, the nitrate at 10 m and the surface temperature had a high sensitivity score (Xboost). In constrast, RF predicted the surface temperature, nitrate concentration at 0 m, and temperature difference between the surface and bottom layers as essential factors for bottom NASC.[134]

ML algorithms are used to identify the clicks generated by the mammals and understand the animals' behavior based on the accelerometer. The five different behaviors (chafing, burst swimming, head shaking, resting, and swimming in a semi-captive setting) of young lemon sharks (*Negaprion brevirostris*) were characterized by a voting ensemble (VE) model.[134] In addition, the model predicts the time of day, tidal phase, and season, which are all important aspects in determining lemon shark feeding and provide insights into their feeding ecology.[135] Deep learning was also used to identify the alien species based on the sound they produced.[135,136] A spiking convolutional neural network (SCNN) effectively generalizes the sound produced by different animals (Sea Audio Data set). SCNN recognized mammals' sounds with high accuracy and recall, whereas the recognition of fish was a little tricky.[136] With online learning algorithms, a new Online Sequential Multilayer Graph Regularized Extreme Learning Machine Autoencoder (MIGRATE_ELM), which has an innovative deep learning algorithm (DELE), was trained with the Sea Audio data set, and showed slightly higher performance than SCNN. In many cases, this algorithm produces equal and slightly higher accuracy than the previous SCNN model. However, it reduces the implementation time by 23% more than the SCNN.[137]

### 3.4. Image-Based Identification and Classification of Marine Macro-organisms.

Similar to classification and behavior characterization using acoustic data sets, ML algorithms were also used to classify and identify fishes, mammals, and other marine animals from underwater images. Classification and identification of marine animals through images also help monitor animals' health conditions and the environment. Recent developments in underwater imaging technology created a massive volume of data, making manual identification a time-consuming and challenging process. Also, the complexity of underwater circumstances such as light, temperature, suspended particles, and pressure influence identification were taken into consideration. Using an automated analyzer supported by ML algorithms is an alternate option.

The real-time detection of fish or animals is a difficult task because of the complexity of underwater video or image data sets. Few studies used ML algorithms to detect underwater fish and other animals. The Sparse Representation-based Classification (SRC-MP) with Eigenfaces and Fisherface (extract features from images) recognize the fishes in the coral reef ecosystem of southern Taiwan.[137] Best recognition and identification rates were observed with SRC-MP with Eigenfaces (81.8% and 96%, respectively). Similarly, the fish from the underwater videos (using Fish Recognition Ground-Truth data set-FRGT) were successfully identified by the linear SVM classifier (accuracy 98.64%) with Spatial Pyramid Pooling (SPP) for features extracted.[138] However, these models use one or various features, and improving the accuracy requires large data sets. The problem of inadequate data was solved by implementing transfer learning and deep learning models.[139,140] The linear SVM classifies the fish with features extracted by the pretrained AlexNet.[139] The FRGT data set AlexNet and linear SVM classifier classify the fishes with 99.45% accuracy (Figure 5).

Similarly, the transfer learning applied to the MobileNet V2 model had high validation accuracy (92.89%) and less computing time than Inception V3 and MobileNet V1 in identifying fishes. Also, the model was 40 M in size, which is suitable for embedding in a device for real-time classification of marine animals from the underwater image.[141] To overcome the limitation of data sets,[141] we created a unique training data set synthetical (realistic simulation of Deep Vision) from images captured from the camera fitted to the trawler system. This data set was used to train the deep neural network, which successfully identifies blue whiting, Atlantic herring, and Atlantic mackerel with an accuracy of 94%. This method of creating synthetic data from the collected data may successfully overcome the shortage of training data.

Also, the difficulties in identifying fish from the blurry ocean images and the lack of training sets were overcome by a CNN model using data augmentation, network simplification, and speeding up the training process. In this model, overfitting was solved by the dropout algorithm, and the parameters inside the network were refined by loss functions.[142] These processes speed up the training process and reduce training loss. Also, the model shows good accuracy and is suitable for embedded systems with autonomous underwater vehicle.[143] Moreover, the combination of human annotation with ML algorithms successfully handled large underwater image data sets to classify mesofauna.[143] Two-step human annotation followed by ML classification was used. The data sets were first annotated by humans and classified by AlexNet. The model shows the inaccuracy in human annotation as a significant factor that affects classification accuracy, where the marking size and false positives show minor influence. Even with advanced deep learning algorithms, the rate of misclassification is high. To reduce the misclassification rate, a species-specific confidence threshold was introduced. A CNN-based framework automatically calculates species-specific confidence threshold value from the training data set (Independent of the data used to train the deep learning algorithm). These threshold values are used in the postprocessing deep learning output, by assigning classification scores for each class and marking a new class as unsure.[144] Applying species-specific threshold values reduces the misclassification rate from 22% to 2.98% in identifying 20 fish species from 13,232 images from coral reef environments.

Besides fish identification, ML algorithms were also used to identify and count whales. The CNN (two-step) successfully identified and counted whales from the image data sources such as satellite and aerial pictures.[145] The first CNN detected the presence of whales in the images, and the second CNN counted the number of whales in those images. The model showed 81% and 94% accuracy in detecting and counting whales from 10 global whale-watching hotspots (Google Earth images data sets), and combining these two CNNs increased the detection rate to 36%. This new tool improves the ongoing efforts in mammal watching and conservation of the vast uncharted regions of the sea. Increasing the availability of satellite and image data sets will lead to better monitoring of endangered mammals.

Exploring deep sea animals by humans is challenging because of the hostile conditions, so identification through images or videos with the aid of ML algorithms is more suitable. However, the identification of animals from the images has difficulties as the underwater images have uneven lamination, noise, and low contrast, which requires some improvements. A modified deep CNN based on region based-CNN (R-CNN) and a modified hypernet method successfully detects and classifies underwater marine organisms.[146] Data sets from a remotely operated vehicle (ROV) (video from a sea cucumber fishing site) and an underwater robot picking contest were used, and the Regional Proposed Network optimized the feature extraction. The CNN model performed well in recalling and detecting organisms, even with a different focus. When the Intersection over Union equals 0.7, the mAP is more than 90%. The model seems suitable for analyzing organisms' real-time detection from a camera installed in an ROV. Similarly, the problem in the deep-sea underwater images, like uneven lamination, noise, and low contrast, was successfully overcome by image enhancement using a combination of two methods, max-RGB and shades of gray and CNN, to solve weak illumination. After preprocessing, scheme two detects and classifies the animals at 50 frames per second detection speed with a mAP of 90%. This ML algorithm is helpful in real-time detecting underwater organisms and can assist underwater robots in avoiding dangerous high-pressure conditions and helping humans understand deep-sea environmental conditions.[147]

A deep neural network along with marine object-based image analysis (MOBIA) efficiently identified the individual organismal distribution and zonation across the CWC Piddington coral mound in Ireland.[148] Two mm high-resolution reef-scale video mosaic and multibeam data from ROV from the CWC Piddington coral mound within the Porcupine Seabight, Ireland Margin, were used for the training. Among the tested models (decision tree, logistic regression, and deep neural network), the deep neural network had higher classification accuracy and recall, which showed that the mound was made up of 12.5% coral rubble, 2% of live corals, and 3.5% of the heterogeneous distribution of sponges in some parts of the mounds. Applying ML provides a baseline to monitor the changes in the mounds. This method can be applied to other habitats to monitor the modifications over a period of time.[149] Likewise, the coral species in the shallow water of the Gulf of Eilat were identified from the underwater images using CNN.[149] The CNN successfully overcomes the difficulties like a coral colony, age, species, species morphology, depth, water current, quality of image, angle of view, etc. With a data set consisting of 11 well-known coral species (5000 underwater images), the model showed an overall accuracy of 80.13% for all 11 species of corals. Among the 11 species, the CNN had high accuracy ranging between 91.5% to 93.5% in identifying *Montipora*, *Lobophyllia*, and *Stylophora*. Future deep learning might be used for real-time monitoring of the effects caused by global climate change on corals in Eilat and other corals around the world. In addition to these studies to analyze large-scale data sets, "DeepFish", was created using ResNet-50.[150] The model trained with a data set contains 40,000 images of fishes (with classification label) from the underwater marine environment in tropical Australia (20 different marine environments). To a note, pretraining and transfer learning improves the accuracy of deep learning algorithms.[151]

**3.5. Identification and Classification of Benthic Fauna.** Megafaunas play an essential role in the functioning of the benthic ecosystem and act as indicators of environmental change. Manual species identification is time-consuming, and most ecological studies frequently neglect this organism size class. Automated image analysis is a possible way to address practical challenges in identifying mesofaunas. However, diverse megafauna populations make such automated approaches difficult. Schoening et al.[151] created an automatic image analysis system called intelligent Screening of underwater Image Sequences (iSIS) to quantify and examine the diverse group of megafauna species. The iSIS had three steps, i.e., feature extraction, training SVM with extracted features, and utilizing human labeled images containing mesofauna taxa. Then the model predicts the possible taxa position and counts the number of taxa in every field of view. The iSIS performed similarly to human experts when the seabed image data set was used (consisting of eight distinct species recorded in the Arctic deep-sea observatory (HAUSGARTEN)). Taxa like *Bathycrinus stalks* and *Kolga hyalina* were well identified by iSIS. Some species of *Elpidia heckeri* (little sea cucumber) remain difficult for both iSIS and human experts. As a result, advancements in computer-assisted benthic ecosystem monitoring might be an alternate method for reducing human time and limitations.[152]

Likewise, the benthic biodiversity was identified by the inception v3 model (TensorFlow) from the underwater images. The model was trained with increasing images (20−1000 images per taxa) and taxa (7−25). The model performed best when 200 images per taxa (0.78 sensitivity, 0.75 precision) and the least number of taxa were used. Even though the model was not an alternative to manual annotation, this technique could be used to classify individual taxa from the images with high precision. This model might help nonexperts study benthic diversity, which leads to an increase in the database for conservation.[152] Also, identifying broad-scale patterns in the benthic faunas is too difficult because the individual benthic surveys could not compare directly. The reliable comparison typically depended on a common set of habitats or a one-off broad-scale spatial survey. Cooper and Barry[153] matched the new benthic fauna survey data with the existing broad-scale cluster group using unsupervised K-mean algorithms. This provides a way to compare individual surveys to identify the macrofaunal clustering patterns. Also, this approach improved the understanding of benthic faunal distribution patterns. An R shiny web application that allows investigators to match habitats with their collected data was also created.

**3.6. Microbiology.** Several researchers have applied ML algorithms to identify or solve the microbe-related problem in the marine system. We have listed a few studies that use ML's potential to solve problems in marine microbiology. High-throughput metagenome sequencing was used to identify the microbial diversity in different environments, from hypersaline sediment[154] to SO waters.[155] Likewise, RF was used to successfully characterize sponges into high microbial abundance (HMA) and low microbial abundance (LMA) groups based on the phylum and class data set. RF model understands the patterns of the host-associated microbiome and, based on the Operational Taxonomic Units (OTUs), predict the status of 135 sponge species without prior knowledge, and divides sponges into four groups (the top two groups consist of HMA = 44 and LMA = 74, respectively). RF proved a valuable tool for addressing host-associated microbial communities' biological questions.[156] Likewise, ML algorithms were successfully applied to differentiate ballast water from the harbor and open sea waters by using 16S rRNA gene sequencing data based on the 16S rDNA OTUs, LefSe, LDA, and ML-predicted sample-specific biomarkers (8 bacteria), which were used in other classification models. With these biomarkers, KNN and RF accurately (80% and 88%, respectively) differentiated the ballast water samples from the harbor and open sea waters samples.[157] Moreover, a strong link between the genome content and ecological niches was predicted by the Gradient boosting (GB) model. About 1961 metagenome-assembled genomes (MAG) were binned from 123 water samples in the Baltic Sea, which belong to 352 species-level clusters corresponding to 1/3 of the metagenome sequences of the prokaryotic size fraction used in the prediction. ML proved that other than phylogenetic signals, genome contents could be used to predict ecological niches.[158] Like biomarker prediction, ML models were used to predict the critical bacterial sub-OTUs (s-OTUs) associated with copepod genera. RF and GB predicted the important bacteriome associated with five different copepod genera viz., *Acartia* spp., *Calanus* spp., *Centropages* sp., *Pleuromamma* spp., and *Temora* spp… The gradient boosting classifiers predicted a total of 50 s-OTUs as important in five copepod genera. Among the predicted s-OTUs, s-OTUs representing the *Acinetobacter johnsonii*, *Phaeobacter*, *Vibrio shilonii*, and Piscirickettsiaceae were reported as important s-OTUs in *Calanus* spp., and the eight s-OTUs representing *Marinobacter, Alteromonas, Desulfovibrio, Limnobacter, Sphingomonas, Methyloversatilis, Enhydrobacter*, and Coriobacteriaceae were predicted as important s-OTUs in *Pleuromamma* spp. for the first time.[159]

Identification of individual microbes requires sophisticated instruments, specific media composition, and time. Also, it is a high risk to culture and identify pathogens related to a biological-risk-related emergency. The recent development of single-cell Raman spectroscopy (scRS) contains a 1000 Raman band, a single-cell fingerprint that represents the cells' inherent phenotype, genotype, and physiological information. However, analyzing scRS is challenging because it requires a sequential process that consumes time. An advanced one-dimensional CNN classification algorithm (1DCNN) proved to be an effective way to analyze scRS data to identify microbes automatically. Along with other ML algorithms like KNN, SVM, PCA−LDA, and 1DCNN accurately classified the microbes from 10 actinomycetes, two nonmarine actinomycetes, and the *E. coli* (reference species) scRS data set. 1DCNN had similar accuracy to other models (∼95%), but the recall rate was higher than other models.[160] Later in the classification of deep-sea microbes using Raman spectra, the addition of progressive growing of generative adaptive nets (PGGAN) enhanced the classification accuracy. PGGAN created a spectral data set similar to actual spectra data acquired from single-cell Raman spectra from the five deep-sea bacteria. The residual network (ResNet) accurately classified bacteria (accuracy of 99.8 ± 0.2%) using the PGGAN data set. The use of PGGAN proved to be an efficient data augmentation method to handle low amounts of data and provides an advantage to analyze the spectrum with a low signal-to-noise ratio. Moreover, the model reduced the requirement of a large data set for training data.[161]

■ AUTHOR INFORMATION

**Corresponding Author**

   Mangesh U. Gauns − *Plankton Laboratory, Biological Oceanography Division, CSIR-National Institute of Oceanography, Dona Paula, Goa 403004, India;*
   orcid.org/0000-0002-4737-9252; Email: gmangesh@nio.org

**Authors**

   Balamurugan Sadaiappan − *Department of Biology, United Arab Emirates University, Al Ain 971, UAE; Plankton Laboratory, Biological Oceanography Division, CSIR-National Institute of Oceanography, Dona Paula, Goa 403004, India*

   Preethiya Balakrishnan − *Faraday-Fleming Laboratory, London W148TL, United Kingdom; University of London, London WC1E 7HU, United Kingdom*

   Vishal C.R. − *Plankton Laboratory, Biological Oceanography Division, CSIR-National Institute of Oceanography, Dona Paula, Goa 403004, India*

   Neethu T. Vijayan − *Plankton Laboratory, Biological Oceanography Division, CSIR-National Institute of Oceanography, Dona Paula, Goa 403004, India*

   Mahendran Subramanian − *Faraday-Fleming Laboratory, London W148TL, United Kingdom; Department of Computing, Imperial College, London SW7 2AZ, United Kingdom*

Complete contact information is available at:
https://pubs.acs.org/10.1021/acsomega.2c06441

**Author Contributions**

B.S., M.S., and M.G. designed and wrote the initial draft. B.S., V.C.R., and P.B. contributed to creating figures. P.B. and N.T.V. contributed to review of the literature and editing. Editing and rewriting were performed by B.S., M.S., and M.G.

**Notes**

The authors declare no competing financial interest.

■ ACKNOWLEDGMENTS

The authors thank the Director, CSIR-NIO, for encouraging this work. B.S., V.C.R., N.T.V., and M.G. received financial assistance from the Council of Scientific & Industrial Research, Government of India, under projects OLP2005 and MLP1802. M.S. is also funded by the Engineering and Physical Sciences Research Council, UK, and Imperial College London (EP/N509486/1:1,979,819). P.B. is funded by Faraday Fleming Laboratory, London, UK. We thank our funders. This is NIO's contribution No 11028. The authors declare that they have no conflict of interest.

■ GLOSSARY

   ML Machine Learning
   AI Artificial Intelligence
   RF Random Forest Classifier
   CNN Convolutional Neural Network
   RL Reinforcement Learning
   GMM Gaussian Mixture Model
   DO Dissolved Oxygen
   SO Southern Ocean
   SOSE Southern Ocean State Estimate
   WOA13 World Ocean Atlas
   M-DJINN Marine-Deep Jointly Informed Neural Network
   DJINN Deep Jointly Informed Neural Network
   EWT Empirical Wavelet Transformation
   RFR Random Forest Regression
   RRF Random Regression Forest
   SVR Support Vector Regression
   OMZs Oxygen Minimum Zones
   NOAA/ESRL National Ocean and Atmospheric Administration Earth System Research Laboratories
   UCYN-A Unicellular Cyanobacteria Group A
   ANN Artificial Neural Network
   MLP Multilayer Perceptron

SVM Support Vector Machine
GPSM Global Predictive Seabed Model
RFRE Random Forest-based Regression Ensemble
KNN K Nearest Neighbors
TOC Total Organic Carbon
LSTM Long Short-Term Memory Neural Network
GPSM Global Predictive Seabed Model
SIPPER Shadow Image Particle Profiling Evaluation Recorder
G-flip Greedy Feature Flip Algorithm
DCNN Deep Convolutional Neural Network
FORABOT FORAminifera roBOT
DNN Deep Neural Network
MKL Multiple Kernel Learning
NA Northern Adriatic Sea
-WA-PE- Weighted Average Prediction Error
BRT Boosted Regression Tree
PERMANOVA Pairwise Permutational Multivariate Analysis of Variance
NPP Net Primary Productivity
BATS Bermuda Atlantic Time-Series Study
GP Genetic Programming
PMID Phytoplankton Microscopic Image Data Set
Chl-a Chlorophyll-a
SPM Suspended Particulate Matter
GOCI Geostationary Ocean Color Imager
OCN Ocean Color Net
GPR Gaussian Process Regression
OC3 Ocean Color algorithm,
C2RCC Case-2 Regional Coast Color
HAB Harmful Algal Bloom
LDA Linear Discriminant Analysis
NN Neural Networks
iSIS intelligent Screening of underwater Image Sequences
ETP Eastern Tropical Pacific
AUV Automated Underwater Vehicle
DLA Deep Learning Algorithms
ROV Remote Operated Vehicle
MOBIA Marine Object-Based Image Analysis
FPS Frames Per Second
MIGRATE_ELM Multilayer Graph Regularized Extreme Learning Machine Autoencoder
VE Voting Ensemble
ISIIS in situ Ichthyoplankton Imaging System
scRS single cell Raman Spectroscopy
LEfSe Linear Discriminant Analysis (LDA) Effect Size
MAG Metagenome-Assembled Genomes
BFGS Broyden−Fletcher−Goldfarb−Shanno
ENN Extended Nearest Neighbor
GRNN General Regression Neural Network
ORELM Outlier Robust Extreme Learning Machine
ELM Extreme Learning Machines combined to obtain the BEGOE model

## ■ REFERENCES